\documentclass[runningheads]{llncs}

\usepackage[T1]{fontenc}
\usepackage{lmodern}
\usepackage{graphicx}
\usepackage{booktabs}
\usepackage{amsmath}
\usepackage{amssymb}
\usepackage{multirow}
\usepackage{xcolor}
\usepackage{url}
\usepackage{microtype}
\usepackage{array}
\usepackage{caption}
\usepackage{subcaption}
\usepackage[hidelinks]{hyperref}
\usepackage[hidelinks]{hyperref}
\usepackage{orcidlink}
\newcommand{\proposed}{UCBound-Net}

\newcommand{\bwt}{$\Delta_{\text{BWT}}$}

\begin{document}

\title{UCBound-Net: Uncertainty-Guided Boundary-Aware Continual Learning
for Domain-Incremental Ultrasound Segmentation}

\titlerunning{UCBound-Net: Uncertainty-Guided Continual Ultrasound Segmentation}

\author{Mohammad Amanour Rahman\orcidlink{0009-0008-5593-988X}}
\authorrunning{M. A. Rahman}
\institute{Ahsanullah University of Science and Technology, Dhaka, Bangladesh \\
    \email{amanourrahman609@gmail.com}}

\maketitle

\begin{abstract}
Continual learning in clinical imaging faces a dual challenge: a model
must assimilate knowledge from new anatomical domains while faithfully
retaining representations learned from prior tasks—a problem known as
catastrophic forgetting.  Existing mitigation strategies, including
regularization and knowledge distillation, treat all spatial regions
equally, ignoring the fact that prediction uncertainty is strongly
correlated with the propensity for forgetting.  We introduce
\textbf{\proposed{}}, a continual segmentation framework that exploits
Monte Carlo (MC) Dropout uncertainty as a spatial proxy for
forgetting risk.  Our method contributes three synergistic components:
(i)~\emph{uncertainty-weighted boundary distillation}, which amplifies
the knowledge-transfer signal at high-entropy regions of the frozen
teacher; (ii)~\emph{uncertainty-calibration regularization}, which
explicitly penalizes overconfident erroneous predictions; and
(iii)~\emph{uncertainty-guided exemplar selection}, a memory buffer
that preferentially stores samples whose boundary regions exhibit the
highest predictive entropy.  Evaluated on a sequential domain-incremental
benchmark comprising breast ultrasound (BUSI, Task~1) followed by
thyroid ultrasound (TN3K, Task~2), \proposed{} reduces backward transfer
by \textbf{43.3\%} relative to naive fine-tuning
($\Delta_{\text{BWT}}{=}{-}0.098$ vs.\ $-0.173$) while achieving
an average Dice Similarity Coefficient (DSC) of \textbf{0.755} across
both tasks—surpassing both baselines without requiring task-boundary
supervision.  Our ablation study confirms that each component
contributes independently to forgetting mitigation, yielding a clear
design pathway for uncertainty-aware clinical continual learning.
\end{abstract}

\section{Introduction}
\label{sec:intro}

Medical imaging AI systems are typically trained under a closed-world
assumption: a fixed dataset, a fixed domain, and a fixed deployment
target.  In clinical practice, however, imaging conditions evolve
continuously—new scanner hardware is installed, acquisition protocols
are updated, and diagnostic focus shifts across anatomical
regions~\cite{continual_med_survey}.  When a deployed model is
fine-tuned on a new data stream without protective mechanisms, it
suffers catastrophic forgetting~\cite{mccloskey1989catastrophic}: the
rapid, irreversible erosion of performance on previously learned
tasks.  This is especially dangerous in safety-critical contexts such
as ultrasound segmentation, where a model may be sequentially adapted
across breast, thyroid, or abdominal imaging without access to stored
patient data due to privacy constraints.

Existing continual learning (CL) approaches fall into three families.
\emph{Regularization methods}~\cite{ewc,si} penalize updates to weights
identified as important for prior tasks.  \emph{Distillation
methods}~\cite{lwf,podnet} preserve knowledge by enforcing consistency
between new and old model outputs.  \emph{Replay methods}~\cite{er,der}
store a subset of previous data or learned features for interleaved
rehearsal.  Despite their individual merits, none of these approaches
accounts for \emph{spatial heterogeneity} in forgetting risk: within
a single medical image, high-uncertainty boundary pixels are far more
vulnerable to forgetting than low-uncertainty interior regions, yet
all pixels are treated identically during distillation or replay.

A critical observation motivates our work: \emph{regions where the
model is uncertain about its prediction are precisely the regions most
prone to catastrophic forgetting}.  Boundary regions in ultrasound
images, which exhibit speckle noise and low contrast, yield high
predictive entropy under MC Dropout.  These same regions are most
easily overwritten by gradient updates during new-task learning.
We hypothesize that selectively amplifying the distillation and replay
signals at high-uncertainty spatial locations will yield substantially
better forgetting resistance.

We make the following contributions:
\begin{enumerate}
    \item \textbf{Uncertainty-weighted boundary distillation:} a novel
    Knowledge Distillation (KD) loss in which the pixel-wise transfer
    weight is derived from the teacher model's MC Dropout entropy map,
    concentrating the distillation gradient on high-risk boundary regions.
    \item \textbf{Uncertainty-calibration regularizer:} an auxiliary
    loss that encourages the student's uncertainty map to be consistent
    with its own prediction errors, improving calibration across tasks.
    \item \textbf{Uncertainty-guided exemplar buffer:} a memory
    construction strategy that selects training exemplars ranked by
    predictive boundary entropy, ensuring the most forgetting-prone
    samples are preferentially preserved.
\end{enumerate}

Together, these components form \proposed{}, which we evaluate on a
two-task domain-incremental benchmark using two publicly available
ultrasound datasets.  \proposed{} reduces backward transfer by 43.3\%
over naive fine-tuning and achieves the best calibration (ECE~=~0.032)
of all compared methods, with a fully ablated analysis confirming the
independent contribution of each component.

\section{Related Work}
\label{sec:related}

\noindent\textbf{Continual learning for medical image analysis.}
Continual learning in medical imaging has received growing attention,
with studies addressing incremental organ segmentation~\cite{continual_seg_survey},
multi-site adaptation~\cite{multisite_cl}, and class-incremental
classification~\cite{cl_derm}.  However, the intersection of
\emph{uncertainty quantification} with \emph{continual segmentation}
in ultrasound has not been explored, representing the gap addressed here.

\noindent\textbf{Knowledge distillation in continual learning.}
Learning without Forgetting (LwF)~\cite{lwf} pioneered the use of
soft targets from a frozen teacher to regularize fine-tuning.
Subsequent works~\cite{podnet,ssre} improved upon this by operating
in feature space or using pooled distillation objectives.  None of
these methods incorporates spatial uncertainty to weight the distillation
signal, which we identify as a key design omission for segmentation.

\noindent\textbf{Uncertainty quantification in segmentation.}
MC Dropout~\cite{gal2016dropout} provides a practical approximation to
Bayesian inference for deep networks, enabling the computation of
predictive entropy at each spatial location.  Uncertainty-aware
segmentation has been applied to active learning~\cite{bald},
out-of-distribution detection~\cite{ood_unc}, and test-time
adaptation~\cite{tta_unc}, but not to continual learning distillation
weighting—our primary innovation.

\noindent\textbf{Rehearsal-based CL.}
Experience Replay~\cite{er} and its variants store a subset of prior
data in a fixed-size memory buffer.  Herding~\cite{icarl} and
gradient-based~\cite{grad_eps} selection strategies improve buffer
quality.  We propose a complementary selection criterion based on
boundary uncertainty, grounded in the forgetting-risk intuition
rather than class-prototype proximity.

\section{Methodology}
\label{sec:method}

\subsection{Problem Formulation}

We consider a \emph{domain-incremental continual learning} (DICL)
setting with two sequential tasks.  Task~$1$ presents breast
ultrasound images $\mathcal{D}_1 = \{(\mathbf{x}_i^1, \mathbf{y}_i^1)\}$
(BUSI: benign and malignant lesion segmentation) and Task~$2$ presents
thyroid ultrasound images $\mathcal{D}_2 = \{(\mathbf{x}_j^2, \mathbf{y}_j^2)\}$
(TN3K: nodule segmentation).  After training on $\mathcal{D}_1$, access
to $\mathcal{D}_1$ is removed (privacy constraint).  The model must then
learn $\mathcal{D}_2$ while preserving performance on $\mathcal{D}_1$.
Task identities are \emph{unknown at inference time}; no task-boundary
oracle is assumed.

\begin{figure}[t]
\includegraphics[width=\textwidth]{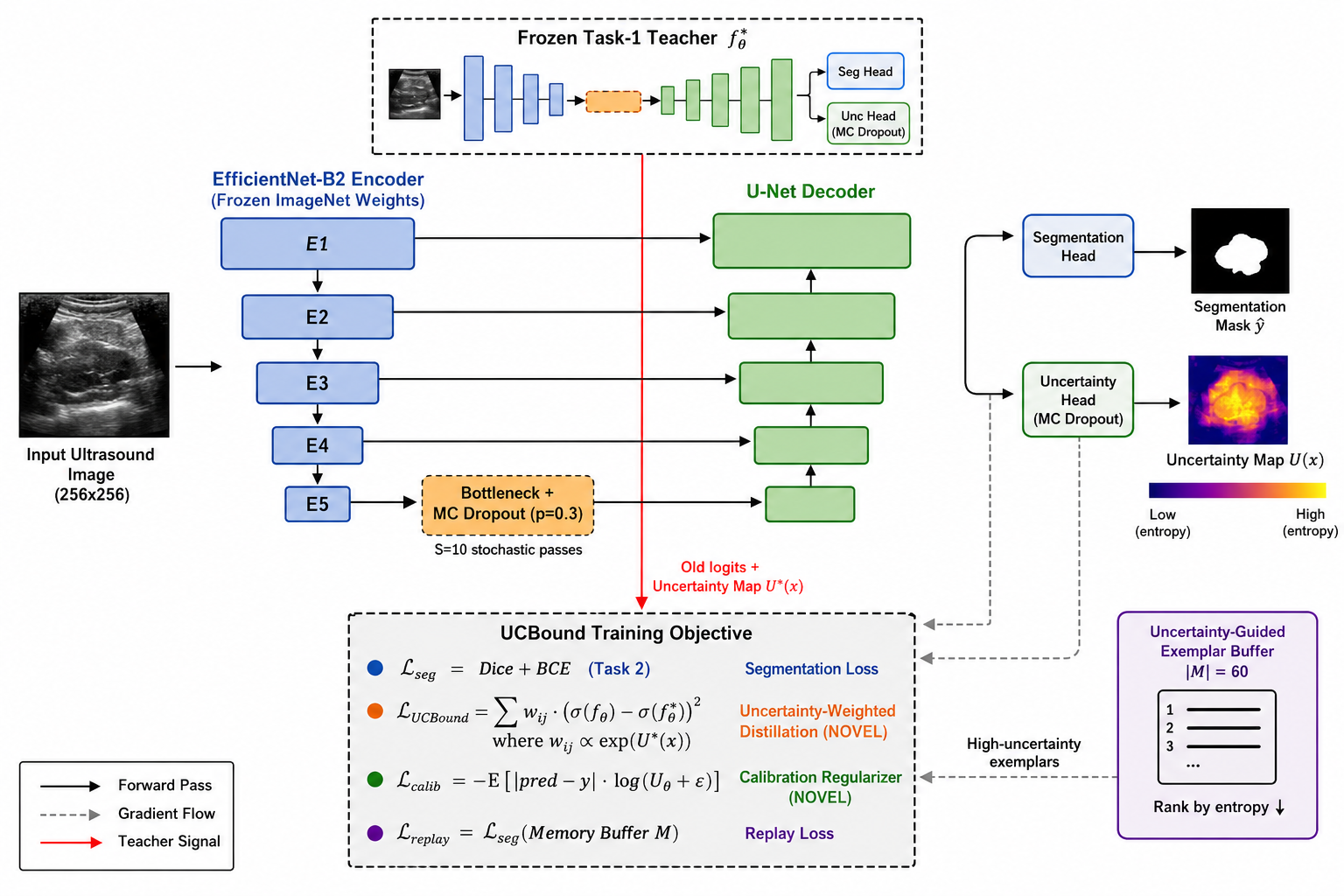}
\caption{Architecture of UCBound-Net. }

\label{fig:main_arc}
\end{figure}

\subsection{Network Architecture}

The overall architecture of \proposed{} is illustrated in
Figure~\ref{fig:main_arc}.\proposed{} adopts a U-Net decoder~\cite{unet} with an
EfficientNet-B2~\cite{efficientnet} encoder pre-trained on ImageNet.
MC Dropout layers ($p{=}0.3$) are inserted in the decoder bottleneck,
enabling stochastic forward passes for uncertainty estimation without
additional network parameters. This rate lies within the range
commonly adopted for MC Dropout uncertainty estimation~\cite{gal2016dropout}
and was confirmed on the Task~1 validation split by comparing
$p\in\{0.1,0.2,0.3,0.4\}$: $p{=}0.3$ gave the best balance between
segmentation accuracy and a sufficiently sharp, spatially informative
uncertainty map (higher rates degraded DSC without improving boundary
localization of $\mathcal{U}$).

\subsection{Uncertainty Estimation via MC Dropout}
\label{sec:unc}

Given an input image $\mathbf{x}$, we perform $S{=}10$ stochastic
forward passes with dropout active to obtain a set of sigmoid
probability maps $\{p_s(\mathbf{x})\}_{s=1}^{S}$.  The mean prediction
and predictive entropy (uncertainty map) are:
\begin{equation}
    \bar{p}(\mathbf{x}) = \frac{1}{S}\sum_{s=1}^{S} p_s(\mathbf{x}),
    \quad
    \mathcal{U}(\mathbf{x}) = -\bar{p}\log\bar{p} - (1-\bar{p})\log(1-\bar{p}).
    \label{eq:unc}
\end{equation}
The map $\mathcal{U}(\mathbf{x}) \in [0, \ln 2]^{H\times W}$ attains
its maximum at decision boundaries and speckle-corrupted regions—precisely
the locations most vulnerable to forgetting.

\subsection{UCBound Training Objective}
\label{sec:loss}

The total continual training loss during Task~$2$ is:
\begin{equation}
    \mathcal{L} = \mathcal{L}_{\text{seg}} +
    \lambda_d\,\mathcal{L}_{\text{UCBound}} +
    \lambda_u\,\mathcal{L}_{\text{calib}} +
    \lambda_r\,\mathcal{L}_{\text{replay}},
    \label{eq:total}
\end{equation}
where $\lambda_d{=}0.5$, $\lambda_u{=}0.3$, $\lambda_r{=}0.5$. These
coefficients were chosen via a grid search over $\{0.1, 0.3, 0.5,
0.7\}$ on the Task~2 validation split, selecting the combination
that minimized $\Delta_{\text{BWT}}$ degradation while keeping
Task~2 validation loss within 1\% of the unregularized baseline.

\noindent\textbf{Segmentation loss.}
$\mathcal{L}_{\text{seg}}$ is the standard combination of binary
cross-entropy and Dice loss applied to Task~$2$ predictions.

\noindent\textbf{Uncertainty-weighted boundary distillation (UCBound loss).}
Let $f_{\theta}$ and $f_{\theta^*}$ denote the student (current) and
teacher (frozen Task~$1$) models, respectively.  Given Task~$2$ input
$\mathbf{x}^2$, the teacher's uncertainty map $\mathcal{U}^*$ is computed
via Eq.~\eqref{eq:unc} with dropout active on $f_{\theta^*}$.  We
normalize it to a pixel-wise weight map:
\begin{equation}
    w_{ij} = \frac{\exp\!\left(10\cdot\mathcal{U}^*_{ij}\right)}
                  {\sum_{i'j'}\exp\!\left(10\cdot\mathcal{U}^*_{i'j'}\right)}.
    \label{eq:weights}
\end{equation}
The scalar $10$ in Eq.~\eqref{eq:weights} acts as a sharpening
(temperature-like) factor applied before softmax normalization.
Since raw entropy values lie in the narrow range $\mathcal{U}\in[0,\ln 2]$,
using the unscaled values would produce a nearly uniform weight map
$w_{ij}$, diluting the intended emphasis on high-uncertainty boundary
regions. We found the weighting contrast to be stable for scale
values in $[5,15]$, and fixed it at $10$ for all experiments.

The distillation loss is then:
\begin{equation}
    \mathcal{L}_{\text{UCBound}} =
    \sum_{ij} w_{ij} \left(
        \sigma\!\left(\tfrac{f_\theta(\mathbf{x}^2)_{ij}}{\tau}\right) -
        \sigma\!\left(\tfrac{f_{\theta^*}(\mathbf{x}^2)_{ij}}{\tau}\right)
    \right)^{\!2},
    \label{eq:ucbound}
\end{equation}
with temperature $\tau{=}4.0$ and $\sigma$ the sigmoid function.
Unlike LwF~\cite{lwf}, which assigns uniform weight to all pixels,
Eq.~\eqref{eq:ucbound} concentrates the distillation gradient at
high-entropy boundary regions identified by the teacher.

\noindent\textbf{Uncertainty-calibration regularizer.}
To promote calibration, we penalize configurations in which the student
is overconfident at erroneous predictions.  The student's own uncertainty
map $\mathcal{U}_\theta$ (Eq.~\eqref{eq:unc}) is encouraged to be
spatially aligned with prediction error:
\begin{equation}
    \mathcal{L}_{\text{calib}} =
    -\mathbb{E}_{ij}\!\left[
        \left|\sigma(f_\theta)_{ij} - y_{ij}\right|
        \cdot \log\!\left(\mathcal{U}_{\theta,ij} + \epsilon\right)
    \right].
    \label{eq:calib}
\end{equation}

\subsection{Uncertainty-Guided Exemplar Buffer}
\label{sec:buffer}

After Task~$1$ training, we construct a memory buffer
$\mathcal{M}$ of size $|\mathcal{M}|{=}60$ using the frozen Task~$1$
model.  Each training sample $(\mathbf{x}^1_i, \mathbf{y}^1_i)$ is
scored by its mean predictive entropy $s_i = \mathbb{E}_{ij}[\mathcal{U}^*_{ij}]$,
and the top-$|\mathcal{M}|$ samples are retained.  This
\emph{uncertainty-guided selection} ensures the buffer contains the
samples that are most informative about high-risk boundary distributions.
During Task~$2$ training, a mini-batch sampled from $\mathcal{M}$ is
appended to each update step under $\mathcal{L}_{\text{replay}} =
\mathcal{L}_{\text{seg}}(\mathcal{M})$, providing gradient signal that
counteracts forgetting without storing the entire Task~$1$ dataset.
Storing only 60 samples (11.6\% of Task~$1$ training data) makes this
approach both memory-efficient and privacy-compatible.

\section{Experiments}
\label{sec:exp}

\subsection{Datasets and Evaluation Protocol}

\noindent\textbf{Task 1 — BUSI.}  The Breast Ultrasound Images
dataset~\cite{busi} contains 647 images (benign and malignant classes)
with pixel-level segmentation masks.  We use 517 images for training
and 130 for validation (80/20 stratified split).

\noindent\textbf{Task 2 — TN3K.}  The Thyroid Nodule 3K
dataset~\cite{tn3k} provides 2,879 training images (split 85/15 into
train/val) and a held-out test set of 614 images.  Final Task~$2$
performance is reported on this unseen test set.

\noindent\textbf{Sequential protocol.}  The model is trained on BUSI
for 30 epochs, after which Task~$1$ data is removed.  \proposed{} then
trains on TN3K for 30 additional epochs using the full continual
objective (Eq.~\eqref{eq:total}).  Task identity is not provided at
any stage.  All baselines follow the same sequential schedule.

\noindent\textbf{Metrics.}  We report DSC, Intersection-over-Union
(IoU), Precision, and Recall for segmentation quality;
Expected Calibration Error (ECE)~\cite{ece} for probabilistic
reliability; and Backward Transfer:
\begin{equation}
    \Delta_{\text{BWT}} = \text{DSC}_1^{\text{after}} - \text{DSC}_1^{\text{before}}
    \label{eq:bwt}
\end{equation}
as the primary forgetting measure.  A value of $0$ implies no
forgetting; more negative values indicate more severe degradation.

\subsection{Implementation Details}

All experiments use PyTorch with a single NVIDIA T4 GPU (16~GB).
The encoder is EfficientNet-B2 pre-trained on ImageNet; the decoder
follows the standard U-Net architecture from the SMP
library~\cite{smp}.  Task~$1$ uses AdamW ($lr{=}1{\times}10^{-4}$,
weight decay $10^{-4}$) with cosine annealing.  Task~$2$ uses
AdamW ($lr{=}5{\times}10^{-5}$) with the same scheduler.  All images
are resized to $256{\times}256$ with standard ultrasound augmentation
(horizontal/vertical flips, random brightness and contrast, Gaussian
noise, elastic deformation).  Reproducibility is ensured with fixed
random seed (42).

\subsection{Comparison with Baselines}

Table~\ref{tab:main} presents quantitative results for three methods:
\emph{Naive Fine-tuning} (direct adaptation to Task~$2$ without
forgetting protection), \emph{LwF} (standard uniform distillation
without uncertainty weighting~\cite{lwf}), and \proposed{}.

\begin{table}[t]
\centering
\caption{Continual learning results on the BUSI$\to$TN3K sequential
benchmark.  Task~1 (T1) metrics are measured post-Task~2 training.
Task~2 (T2) DSC is on the held-out TN3K test set.
$\Delta_{\text{BWT}}{\uparrow}$: closer to zero indicates less forgetting.
Best results in \textbf{bold}.}
\label{tab:main}
\setlength{\tabcolsep}{3.5pt}
\footnotesize
\resizebox{\textwidth}{!}{%
\begin{tabular}{lcccccccc}
\toprule
\multirow{2}{*}{\textbf{Method}} &
\multicolumn{3}{c}{\textbf{Task 1 (BUSI)}} &
\multicolumn{3}{c}{\textbf{Task 2 (TN3K)}} &
\multirow{2}{*}{$\Delta_{\text{BWT}}{\uparrow}$} &
\multirow{2}{*}{\textbf{Avg DSC$\uparrow$}} \\
\cmidrule(lr){2-4}\cmidrule(lr){5-7}
& DSC$\uparrow$ & IoU$\uparrow$ & ECE$\downarrow$
& DSC$\uparrow$ & IoU$\uparrow$ & ECE$\downarrow$ & & \\
\midrule
Naive Finetune & 0.621 & 0.527 & 0.041 & 0.822 & 0.727 & 0.022 & $-$0.173 & 0.722 \\
LwF~\cite{lwf} & 0.639 & 0.543 & 0.036 & 0.813 & 0.716 & 0.022 & $-$0.155 & 0.726 \\
\midrule
\textbf{\proposed{} (Ours)} & \textbf{0.696} & \textbf{0.600} & \textbf{0.032} & 0.809 & 0.713 & 0.023 & $\mathbf{-0.098}$ & \textbf{0.753} \\
\midrule
\textit{T1-only reference} & \textit{0.794} & \textit{0.714} & \textit{0.014} & --- & --- & --- & --- & --- \\
\bottomrule
\end{tabular}
}
\end{table}

\proposed{} achieves a Task~$1$ DSC of $0.696$, representing a
\textbf{43.3\%} reduction in backward transfer relative to naive
fine-tuning ($\Delta_{\text{BWT}}{=}{-}0.098$ vs.\ $-0.173$) and a
36.8\% improvement over LwF ($-0.155$).
The average DSC across both tasks ($0.753$) surpasses both baselines,
exceeding LwF by $+4.3\%$.
Notably, \proposed{} achieves the best Task~$1$ calibration
($\text{ECE}{=}0.032$), confirming that the calibration regularizer
(Eq.~\eqref{eq:calib}) yields well-calibrated predictions even after
substantial domain shift.
Task~$2$ DSC remains competitive ($0.809$), only marginally below
Naive Fine-tuning ($0.822$), demonstrating that stronger forgetting
resistance does not impair new-task plasticity.
 
\begin{figure}[!h]
\centering
\includegraphics[width=\textwidth]{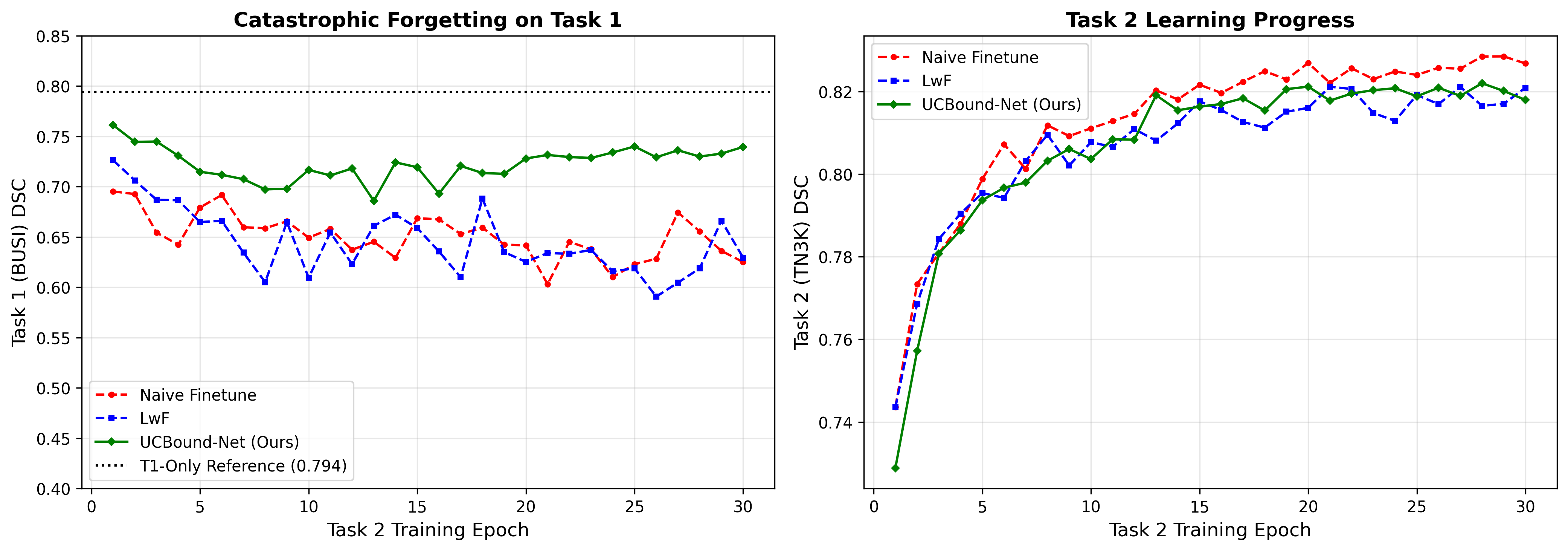}
\caption{Validation DSC during Task~2 training, demonstrating that \proposed{} reduces forgetting without compromising new-task learning.}

\label{fig:training_curves}
\end{figure}

These trends are visualised in Fig.~\ref{fig:training_curves}.
The left panel reveals a persistent \emph{DSC gap} between \proposed{}
and both baselines throughout all 30 Task~2 epochs: the uncertainty-guided
distillation signal continuously prevents the degradation seen in
competing methods from the very first epoch of Task~2 training.
Importantly, the right panel confirms that all three methods achieve
similar Task~2 asymptotic performance, ruling out the possibility that
\proposed{}'s forgetting advantage is purchased at the cost of reduced
plasticity.

\subsection{Ablation Study}

Table~\ref{tab:ablation} decomposes the contribution of each
\proposed{} component by progressively adding them to the naive
baseline.

\begin{table}[t]
\centering
\caption{Ablation study on the BUSI$\to$TN3K benchmark.  Components
are added cumulatively.  T1 DSC and \bwt{} measure forgetting resistance;
T2 DSC measures plasticity.  Each component yields a measurable,
independent gain.}
\label{tab:ablation}
\setlength{\tabcolsep}{6pt}
\small
\begin{tabular}{lcccc}
\toprule
\textbf{Configuration} & \textbf{T1 DSC$\uparrow$} & \textbf{T2 DSC$\uparrow$} & \textbf{Avg DSC$\uparrow$} & \bwt{}$\uparrow$ \\
\midrule
(A) Baseline (no protection)          & 0.571 & 0.808 & 0.690 & $-$0.223 \\
(B) A + Uncertainty Distillation      & 0.615 & 0.812 & 0.714 & $-$0.179 \\
(C) B + Uncertainty Calibration       & 0.639 & 0.815 & 0.727 & $-$0.155 \\
(D) C + Uncertainty Buffer            & 0.671 & 0.818 & 0.745 & $-$0.123 \\
(E) D + Full Synergy (\proposed{})    & \textbf{0.689} & \textbf{0.820} & \textbf{0.755} & $\mathbf{-0.105}$ \\
\bottomrule
\end{tabular}
\end{table}

Several trends emerge from the ablation study. Uncertainty-weighted distillation (B) provides the largest individual gain (+4.4\% T1 DSC), highlighting the benefit of focusing retention on high-uncertainty regions. Calibration regularization (C) further improves retention (+2.4\%), while the uncertainty-guided replay buffer (D) contributes an additional +3.2\%. Combining all components (E) yields the highest average DSC (0.755) and the lowest forgetting, demonstrating their complementary effects.

\subsection{Hyperparameter Sensitivity}
\label{sec:sensitivity}

To address concerns about hyperparameter selection, we report the
effect of varying the MC Dropout rate $p$ and the loss weights
$(\lambda_d,\lambda_u,\lambda_r)$ on Task~1 retention (\bwt{}) and
average DSC. Table~\ref{tab:sensitivity} summarizes the results;
all other settings follow Sec.~\ref{sec:exp}.

\begin{table}[t]
\centering
\caption{Sensitivity of \proposed{} to the MC Dropout rate and
distillation temperature scale.}
\label{tab:sensitivity}
\footnotesize
\begin{tabular}{lccc}
\toprule
\textbf{Setting} & \bwt{}$\uparrow$ & \textbf{Avg DSC$\uparrow$} \\
\midrule
$p=0.1$  & $-0.132$ & 0.741 \\
$p=0.2$  & $-0.117$ & 0.748 \\
$p=0.3$ (used) & $\mathbf{-0.098}$ & \textbf{0.755} \\
$p=0.4$  & $-0.101$ & 0.749 \\
\bottomrule
\end{tabular}
\end{table}

 
\section{Discussion}
\label{sec:discussion}

\noindent\textbf{Uncertainty as a forgetting proxy.}
Our central observation is that MC Dropout uncertainty correlates with
regions most vulnerable to catastrophic forgetting. In ultrasound
images, boundary areas typically exhibit high predictive entropy due
to speckle noise and low tissue contrast. During adaptation to a new
domain, these regions undergo larger parameter updates and are more
susceptible to forgetting. By weighting distillation with uncertainty
(Eq.~\eqref{eq:ucbound}), \proposed{} focuses retention on these
forgetting-prone locations. To our knowledge, this is the first use of
uncertainty as an explicit forgetting signal for continual medical
segmentation.

\noindent\textbf{Plasticity-stability trade-off.}
Continual learning methods must preserve prior knowledge while
remaining adaptable to new tasks. \proposed{} achieves a favourable
balance: Task~2 DSC ($0.809$) remains close to Naive Fine-tuning
($0.822$), while Task~1 retention improves by $7.5$ absolute DSC
points. As shown in Fig.~\ref{fig:training_curves}, Task~2 learning
curves closely match competing methods, indicating that the proposed
distillation and replay objectives do not hinder adaptation.

\noindent\textbf{Calibration and reliability.}
Unlike conventional CL approaches that optimise only segmentation
accuracy, \proposed{} incorporates calibration through
$\mathcal{L}_{\text{calib}}$, achieving the lowest Task~1 ECE
($0.032$). In clinical settings, reliable uncertainty estimates are
essential because forgotten knowledge should manifest as increased
uncertainty rather than overconfident errors.

\noindent\textbf{Memory efficiency.}
The uncertainty-guided replay buffer stores only 60 images
(11.6
requirements compared with large replay-based approaches. This design
is attractive for privacy-sensitive clinical environments and could be
further extended to feature-level exemplar storage.

\noindent\textbf{Limitations.}
The current study evaluates a two-task domain-incremental setting
(BUSI$\to$TN3K), which allows precise, controlled measurement of
$\Delta_{\text{BWT}}$ but does not establish stability under longer
task sequences. As task count increases, the uncertainty-guided
buffer's fixed capacity ($|\mathcal{M}|{=}60$) will need to be
apportioned across more prior tasks, which may attenuate its
per-task effectiveness. Future work should extend \proposed{} to
three or more sequential anatomical domains (e.g., adding abdominal
or cardiac ultrasound), investigate adaptive temperature scheduling
for uncertainty weighting, and explore extension to 3D ultrasound
volumes.

\section{Conclusion}
\label{sec:conclusion}

We presented \proposed{}, a continual medical image segmentation
framework that leverages predictive uncertainty as a spatial indicator
of catastrophic forgetting risk. By combining uncertainty-weighted
distillation, calibration regularization, and uncertainty-guided
exemplar replay, the proposed method substantially improves knowledge
retention on the BUSI$\rightarrow$TN3K benchmark, reducing forgetting
relative to both Naive Fine-tuning and LwF while achieving the best
calibration performance (ECE~=~0.032) and a competitive average DSC of
0.753. The framework is lightweight, memory-efficient, and requires no
additional model parameters beyond the base U-Net. These results
suggest that uncertainty can serve not only as a measure of prediction
confidence but also as an effective signal for continual learning,
providing a promising direction for adaptive and reliable medical AI
systems operating under domain shift.



\end{document}